\pdfoutput=1

\documentclass{cup-hpl}
\usepackage{multirow} 
\usepackage{makecell}
\usepackage{lipsum}

\begin{document}

\newtheorem{theorem}{Theorem}

\shorttitle{Segmentation of Laser-induced Damage}   
\shortauthor{Y.Han et al.}

\title{CG-fusion CAM: Online segmentation of laser-induced damage on large-aperture optics}

\author[1,2]{Yueyue Han}
\author[1]{Yingyan Huang}
\author[1]{Hangcheng Dong}
\author[1]{Fengdong Chen\corresp{Building 2d, Science Park, Harbin Institute of Technology.\email{chenfd@hit.edu.cn}}}
\author[2]{Fa Zeng}
\author[2]{Zhitao Peng}
\author[2]{Qihua Zhu}
\author[1]{Guodong Liu\corresp{Building 2d, Science Park, Harbin Institute of Technology.
                       \email{lgd@hit.edu.cn }}}

\address[1]{School of Instrumentation Science and Engineering, Harbin Institute of Technology, Harbin 150006, China}
\address[2]{Research Center of Laser Fusion, China Academy of Engineering Physics, Mianyang 621900, China}

\begin{abstract}
Online segmentation of laser-induced damage on large-aperture optics in high-power laser facilities is challenged by complicated damage morphology, uneven illumination and stray light interference. Fully supervised semantic segmentation algorithms have achieved state-of-the-art performance, but rely on plenty of pixel-level labels, which are time-consuming and labor-consuming to produce. LayerCAM, an advanced weakly supervised semantic segmentation algorithm, can generate pixel-accurate results using only image-level labels, but its scattered and partially under-activated class activation regions degrade segmentation performance. In this paper, we propose a weakly supervised semantic segmentation method with Continuous Gradient CAM and its nonlinear multi-scale fusion (CG-fusion CAM). The method redesigns the way of back-propagating gradients and non-linearly activates the multi-scale fused heatmaps to generate more fine-grained class activation maps with appropriate activation degree for different sizes of damage sites. Experiments on our dataset show that the proposed method can achieve segmentation performance comparable to that of fully supervised algorithms.
\end{abstract}


\maketitle
\section{Introduction}
In inertial confinement fusion (ICF) experiments\cite{bib1}, high-power laser irradiation can cause laser-induced damage (LID)\cite{bib2} on the surface of the final optics assembly (FOA)\cite{bib3,bib4}. Online detection of the damage status of optics is essential for the safe and efficient operation of the ICF facilities. Therefore, the National Ignition Facility (NIF)\cite{bib5,bib6}, the Laser Megajoule (LMJ) in France\cite{bib7,bib8}, and the laser facility at the China Academy of Engineering Physics (CAEP)\cite{bib9,bib10} have successively developed their final optics damage inspection (FODI) systems\cite{bib11,bib12,bib13} to capture damage images of optics online.After the ICF experiments, the imaging system is fed into the center of the target chamber by the support positioning system and captures images of the optics using dark-field imaging with edge illumination. Figure 1 shows the methodology.

\begin{figure}[ht]
\centering
\includegraphics[scale=0.35]{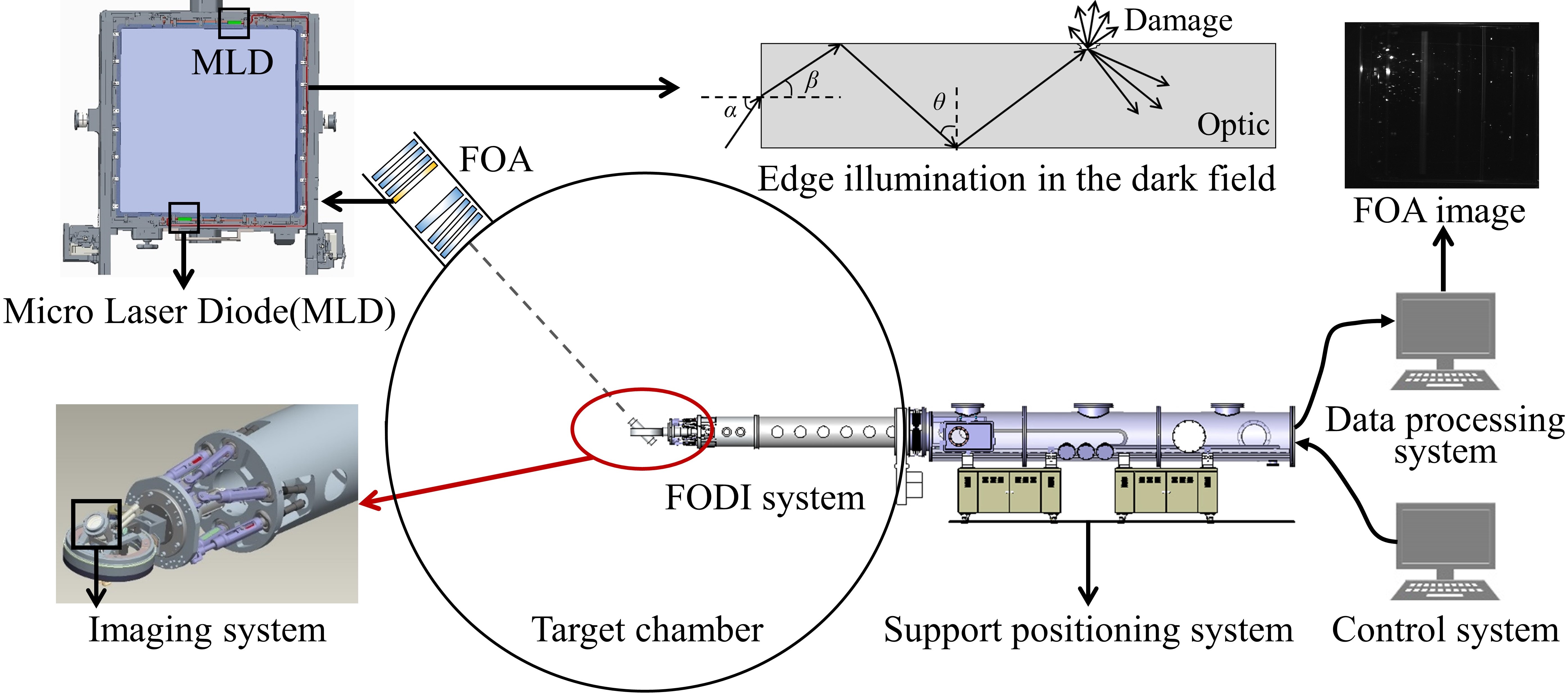}
\caption{Schematic diagram of the methodology in capturing FOA images online by FODI.}
\end{figure}

Dark-field imaging of damage with edge illumination appear as bright spots on a dark background. The damage status of the optics is assessed by locating and segmenting these bright spots online. Complex factors such as large difference in damage size, uneven illumination and stray light interference\cite{bib14} make damage image segmentation more challenging. A typical FOA image is shown in Figure 2.

\begin{figure}[ht]
\centering
\includegraphics[scale=0.38]{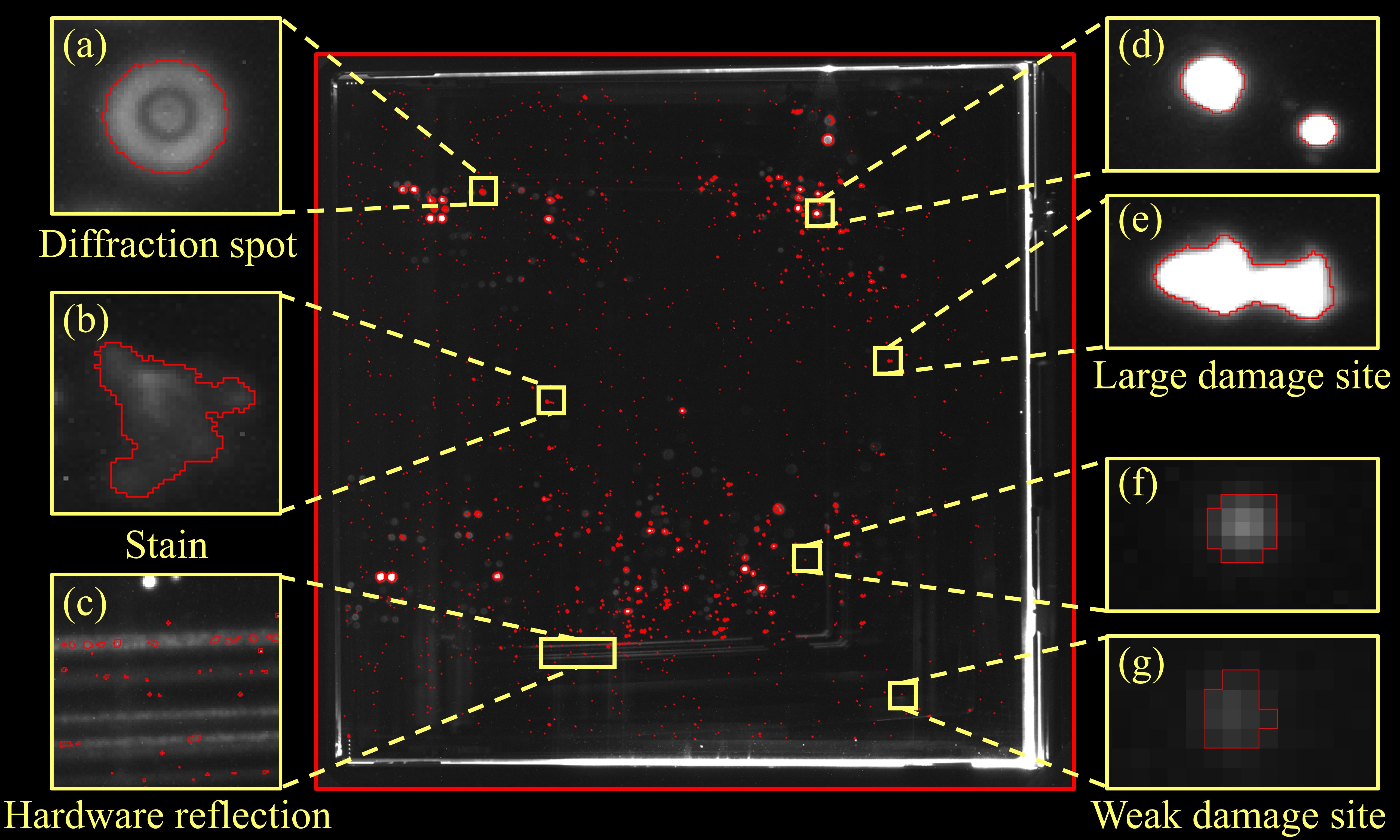}
\caption{An example of an FOA image. (a-c) are images of stray light interference. (d-e) are images of large damage sites. (f-g) are images of weak damage sites.}
\end{figure}

Lawrence Livermore National Laboratory (LLNL) proposed a local area signal-to-noise ratio (LASNR) algorithm\cite{bib15}. This algorithm highlights weak damage signals from its local neighborhood and has a high detection recall. However, its detection accuracy and robustness are largely limited by factors such as changing illumination conditions, stray light interference and noise.

Semantic segmentation algorithms based on deep convolutional neural networks do not require custom-built parameters for the above multi-factor interference scenes, can automatically extract effective damage features and robustly segment damage sites. X. Chu et al. of CAEP constructed a fully convolutional network with a U-shaped architecture (U-Net)\cite{bib16}. Through fully supervised training, this model achieves higher damage detection accuracy than conventional methods. However, producing large quantities of pixel-level labels requires specialist knowledge and a great deal of time and effort.

\begin{figure}[ht]
\centering
\includegraphics[scale=0.55]{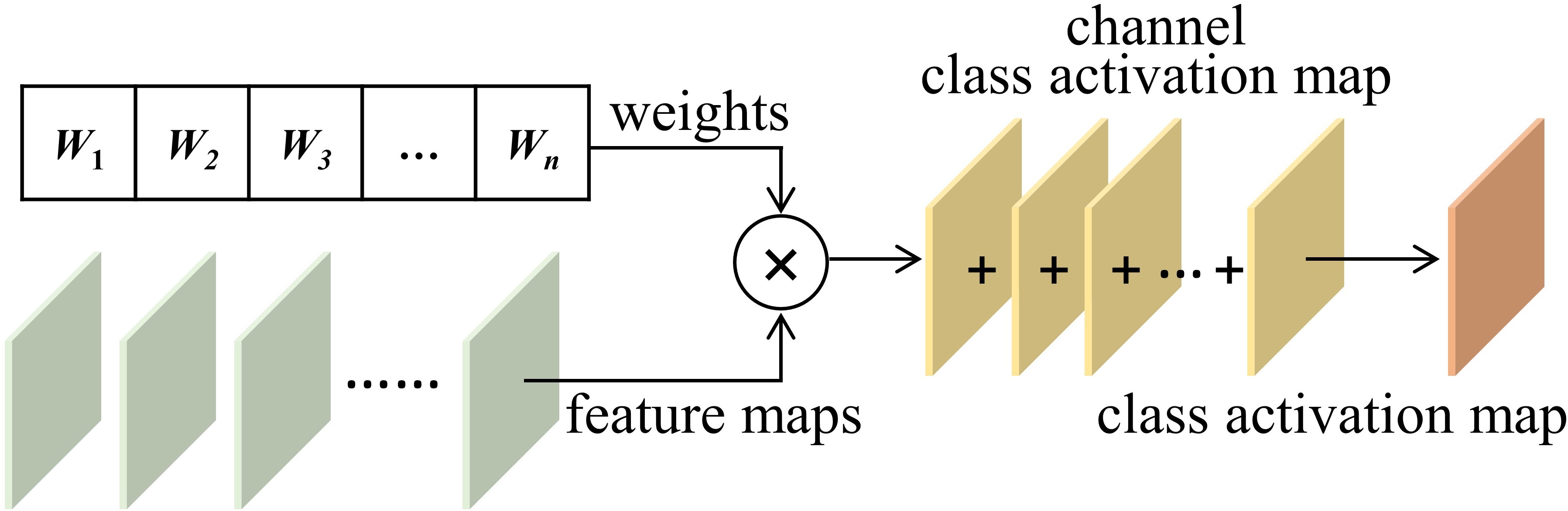}
\caption{The process of class activation mapping.}
\end{figure}

Weakly supervised learning methods can reduce the cost of manual annotation. Currently, state-of-the-art weakly supervised semantic segmentation algorithms are based on class activation maps, the general procedures of these methods are shown in Figure 3. Relying only on image-level labels, deep learning models can generate pixel-level segmentation results. B. Zhou et al.\cite{bib17} first proposed Class Activation Mapping (CAM) to achieve target segmentation by visualizing feature points that play an important role in target classification. However, it is inconvenient to modify the network structure and retrain for CAM in practical applications. Later, Gradient-weighted Class Activation Mapping (Grad-CAM) and its variants\cite{bib18,bib19,bib20} were proposed. They generate class activation maps using the average gradients of the target class score with respect to the feature maps of the final convolutional layer as class activation weights (Global weight CAM). Limited by the spatial resolution of the final convolutional layer, Grad-CAM can only roughly locate the targets. Therefore, P .T. Jiang et al.\cite{bib21} proposed Layer Class Activation Mapping (LayerCAM). They use pixel-level weights to generate reliable class activation maps for each stage and combine them to obtain fine-grained segmentation results (pixel-level weight CAM). Figure 4 shows the typical results of Grad-CAM and LayerCAM on FOA images.

\begin{figure}[ht]
\centering
\includegraphics[scale=0.44]{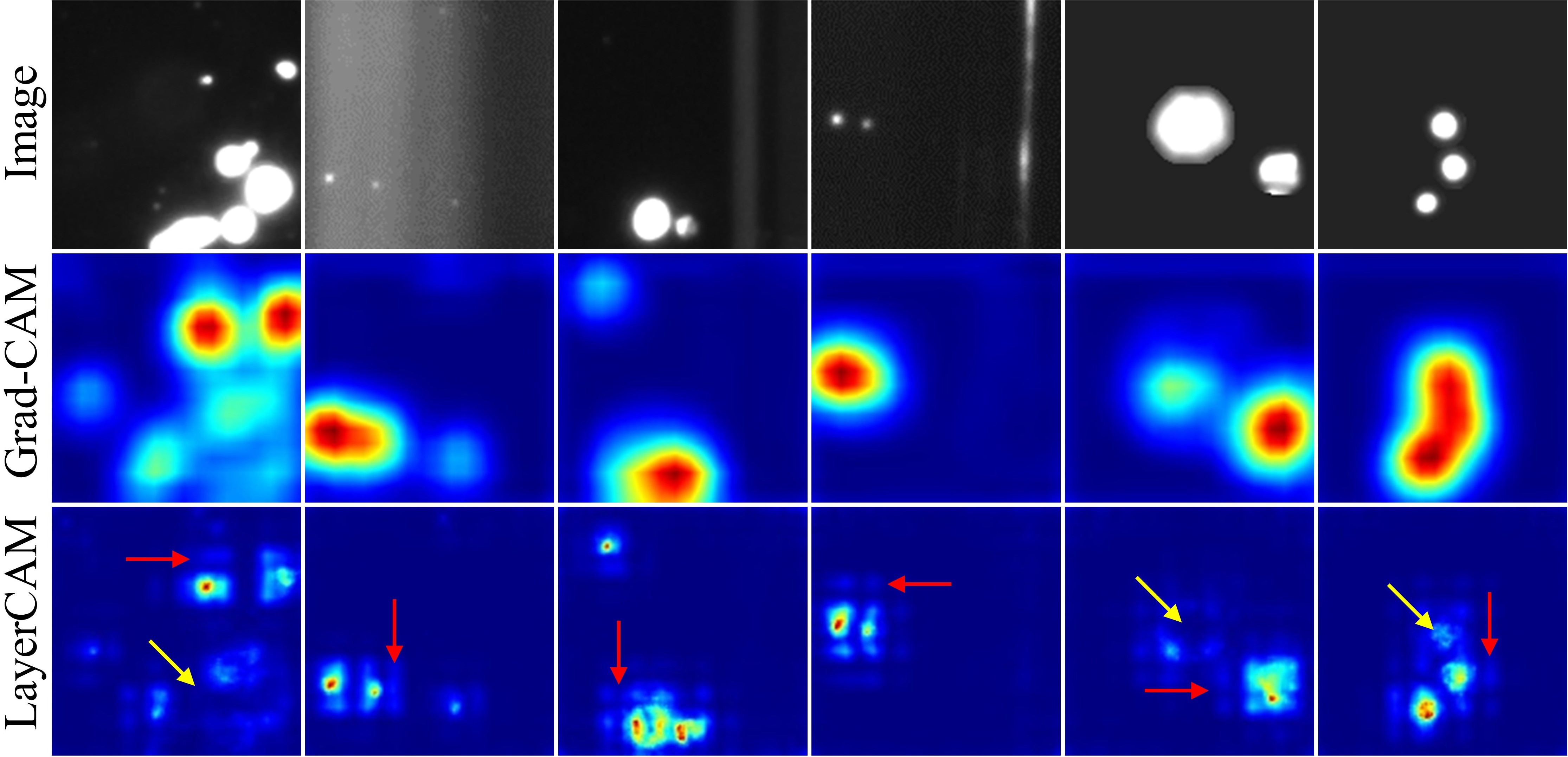}
\caption{Examples of class activation maps generated by Grad-CAM and LayerCAM on FOA images. The red arrows point to scattered activated regions. The yellow arrows point to under-activated regions.}
\end{figure}

However, we found two problems with LayerCAM. One is that the discontinuous pixel-level class activation weights generate scattered class activation regions, resulting in a single damage object being segmented into multiple objects. The other is that large damage sites are under-activated in the class activation maps from shallow layers, leading to degradation in segmentation accuracy or even missed detections. As shown in Figure 4.

To solve the above problems, we propose CG-fusion CAM to generate more fine-grained class activation maps with appropriate activation degree, enabling efficient and accurate detection and segmentation of damage sites with large size differences. 

\begin{figure*}[ht]
\centering
\includegraphics[scale=0.52]{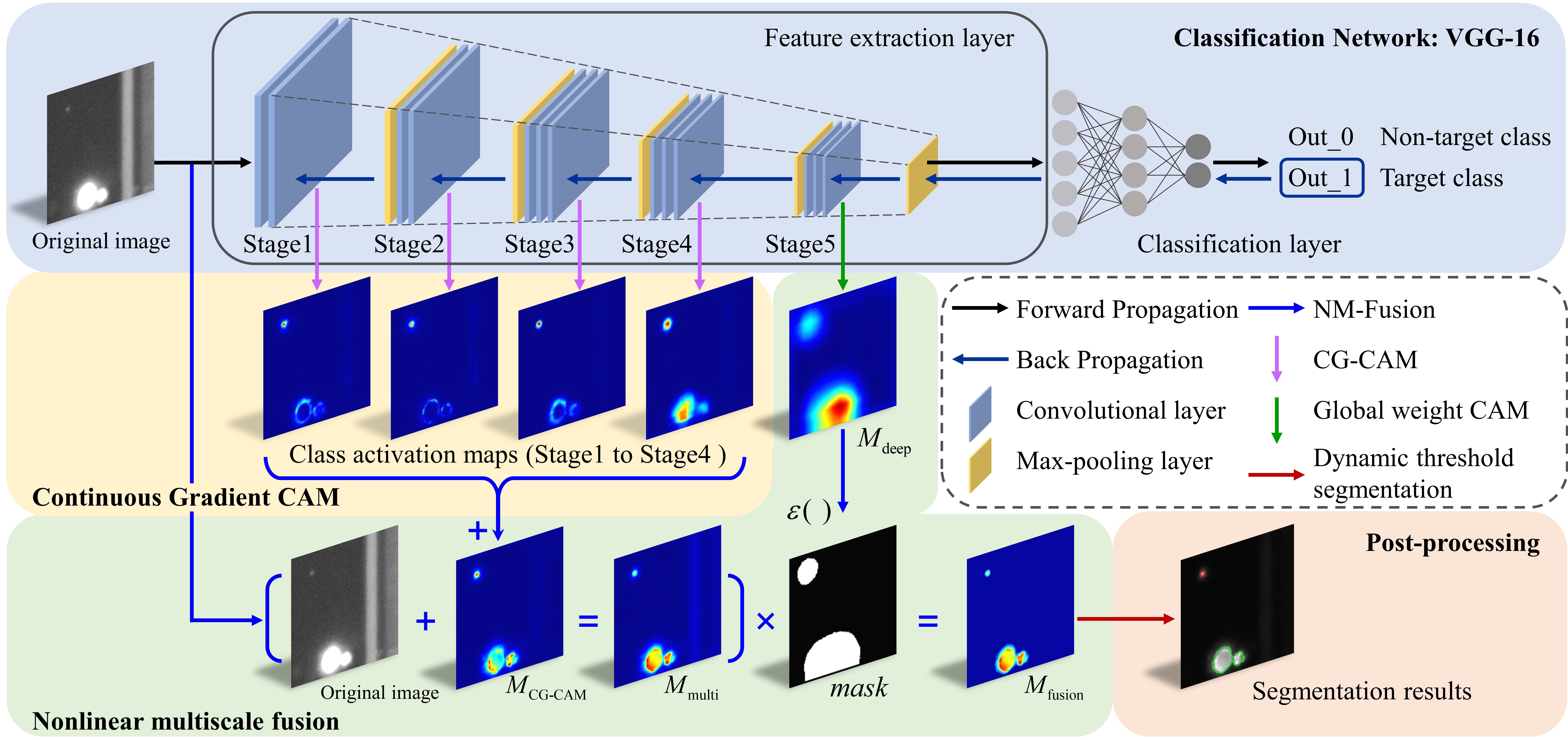}
\caption{The pipeline of CG-fusion CAM.}
\end{figure*}

\section{Methodology}
Our method CG-fusion CAM includes four parts: classification network, Continuous Gradient CAM(CG-CAM), nonlinear multi-scale fusion (NM-Fusion) and post-processing. The overall pipeline is shown in Figure 5. The classification network is used to extract image features and determine their categories. CG-CAM is used to generate more fine-grained class activation maps with continuous activated regions. Unlike the gradient-based CAM algorithms above, we propose a new way of back-propagating gradients. It distributes the gradient of the feature point in the low-resolution layer equally to each feature point (within the same pooling kernel) in the forward (taking the direction of forward propagation as positive) high-resolution layer, thus preserving the fine-grained information lost in down-sampling. NM-Fusion is used to further enhance the effect of class activation in CG-CAM. we propose an algorithm to non-linearly activate multi-scale fused heatmaps. On the basis of improving the fine granularity of class activation maps through linearly fusing multi-scale heatmaps, the original image is used to compensate for the under-activation of large targets based on the complementarity of the target gray values. Subsequently, the high-level semantic information from deep layers is used to non-linearly suppress the stray light interference introduced by the original images. Post-processing is used to segment class activation maps. We choose the dynamic threshold segmentation algorithm to obtain high-precision target regions.

\subsection{Classification network}
The selection of the classification network has an important impact on the effectiveness of feature extraction, the accuracy of the class activation weights and the localization performance of the class activation maps. Therefore, we choose VGG-16 as our classification network with excellent classification ability, simple structure, fast training speed and easy deployment\cite{bib22}. VGG-16 contains 2 parts: a feature extraction layer and a classification layer. Among them, the feature extraction layer contains 5 stages. Each stage consists of several convolutional layers and a max-pooling layer. Each convolutional layer contains multiple channel outputs, also known as feature maps.

\subsection{Continuous Gradient CAM}
\paragraph{Analysis} Inspired by LayerCAM, we use the class activation maps with higher spatial resolution generated from the shallow layers to increase the fine-grained information of targets. However, the typical phenomenon of scattered class activation regions is present at each stage of VGG-16, as shown in Figure 6. We further decompose the class activation map for each channel of each stage into its feature map and gradient map and find that the discontinuity of the gradients associated with the target regions leads to the scattered class activation regions. Figure 6 randomly shows the partial channel results of Stage5.

\begin{figure*}[ht]
\centering
\includegraphics[scale=0.54]{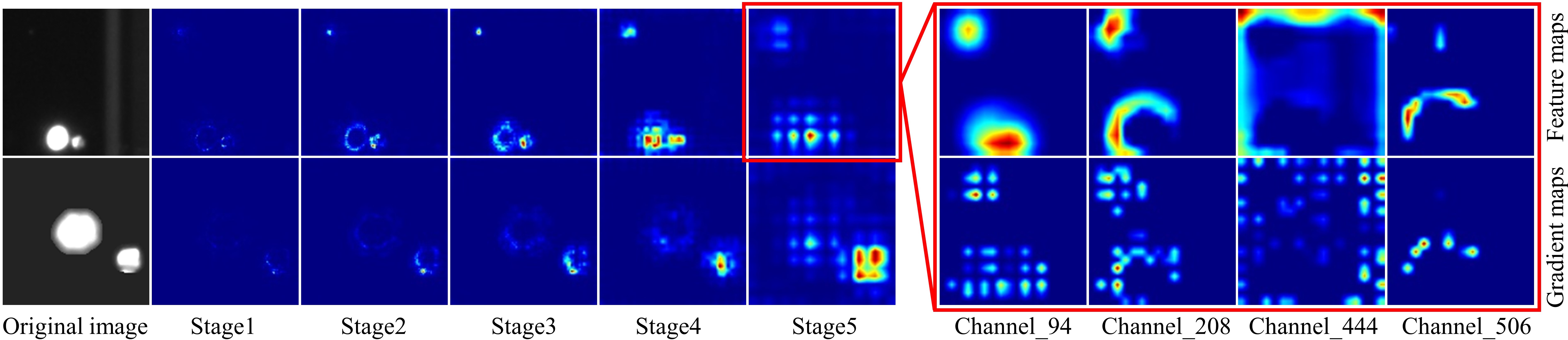}
\caption{The class activation maps of LayerCAM from different stages. The red box shows the feature and gradient maps of some channels from Stage5.}
\end{figure*}

\begin{figure}[ht]
\centering
\includegraphics[scale=0.30]{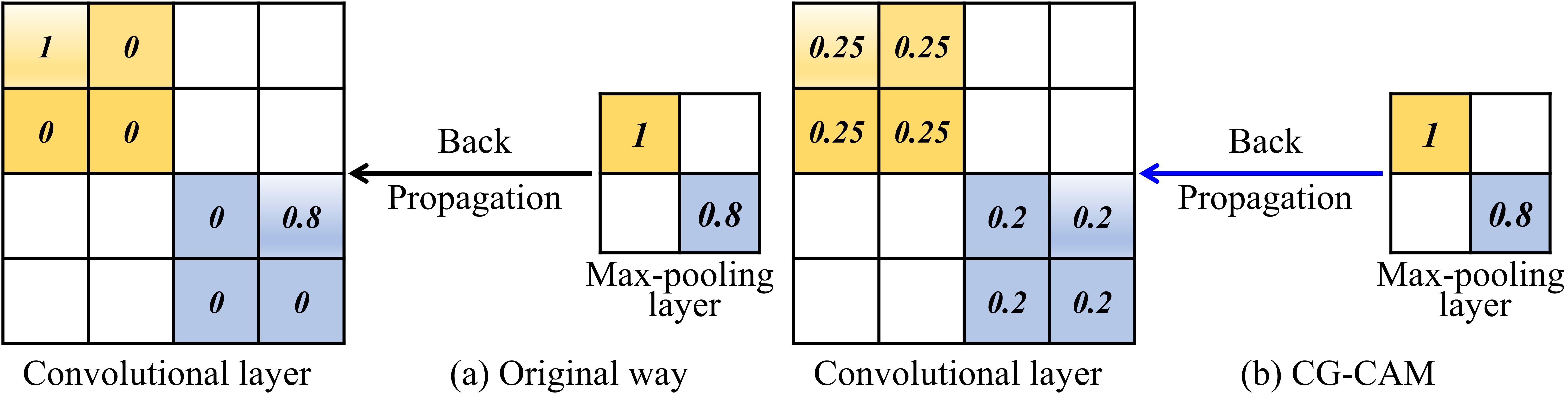}
\caption{The original and CG-CAM way of back-propagating gradients from the max-pooling layer to the convolution layer.}
\end{figure}

According to the chain rule\cite{bib23}, it is known that max-pooling layers are the cause of scattered gradients. This is because in forward propagation, max-pooling layers only retain the largest feature value in each kernel to compress features and reduce computation. In back propagation, each max-pooling layer passes its gradient to the maximum feature value (within the same pooling kernel) in the forward convolution layer, but the non-maximum features have no gradients and are assigned with zero. Figure 7(a) illustrates the way of back-propagating gradients from the max-pooling layer to the convolution layer.

The activation maps generated in the above way do not contain much more fine-grained information. They are the results of up-sampling the low-resolution class activation maps by interpolating zeros. To effectively preserve more fine-grained features, CG-CAM proposed in this paper reassigns reliable gradients for them.

\paragraph{Algorithm} Specifically, in back propagation, CG-CAM extracts the gradients of the last convolutional layer (the forward layer of the max-pooling layer) at each stage by the hook operation\cite{bib24}. Then, the average pooling operation and the up-sampling operation distribute the gradient equally to each feature point within the same pooling kernel, restoring the continuity of the scattered gradients, as shown in Figure 7(b). CG-CAM uses the modified gradients as pixel-level weights to activate corresponding feature points. Formally, the weight $w_{i j}^{k c}$ of the spatial position ($i$, $j$) in the k-th channel feature map of a certain convolutional layer to the target class c can be written as

\begin{equation}
w_{i j}^{k c}=\text {upsample }\left(\text { avgpool2d}\left(g_{i j}^{k c}\right)\right),
\end{equation}
Where $\text {upsample}$ is the up-sampling function, $\text{avg\_pool2d}$ is the average pooling function, $g_{i j}^{k c}$ is the gradient of the predicted score $y^{c}$ (before softmax) of the target class c with respect to the feature map $A_{i j}^{k}$. Its formula is as follows:

\begin{equation}
g_{i j}^{k c}=\frac{\partial y^{c}}{\partial A_{i j}^{k}},
\end{equation}
Where $A_{i j}^{k}$ is the feature value of the spatial position (i,j) in the k-th channel of a certain convolutional layer. The class activation map $M^{c}$ of CG-CAM is calculated as follows:

\begin{equation}
M^{c}=\operatorname{ReLU}\left(\sum_{k} w_{i j}^{k c} \cdot A_{i j}^{k}\right)=\operatorname{ReLU}\left(\sum_{k} \hat{A}^{k}\right),
\end{equation}
Where $\hat{A}^{k}$ is the class activation map of the k-th channel. Linearly summing all channels' $\hat{A}^{k}$ to obtain the $M^{c}$ of a certain convolutional layer. We also apply the ReLU operation to remove the effect of negative gradients.

CG-CAM solves the problem of scattered class activation regions caused by the loss of fine-grained information gradients in LayerCAM. Figure 8 shows the visualization results of the gradients before and after the application of CG-CAM from randomly selected channels from each stage. The gradients generated by CG-CAM are continuous and smooth compared to the original scattered gradients. This allows more semantic regions of the targets to be activated. Comparative experiments demonstrate that CG-CAM can effectively preserve high-resolution features and significantly improve the quality of class activation maps.

\begin{figure*}[ht]
\centering
\includegraphics[scale=0.53]{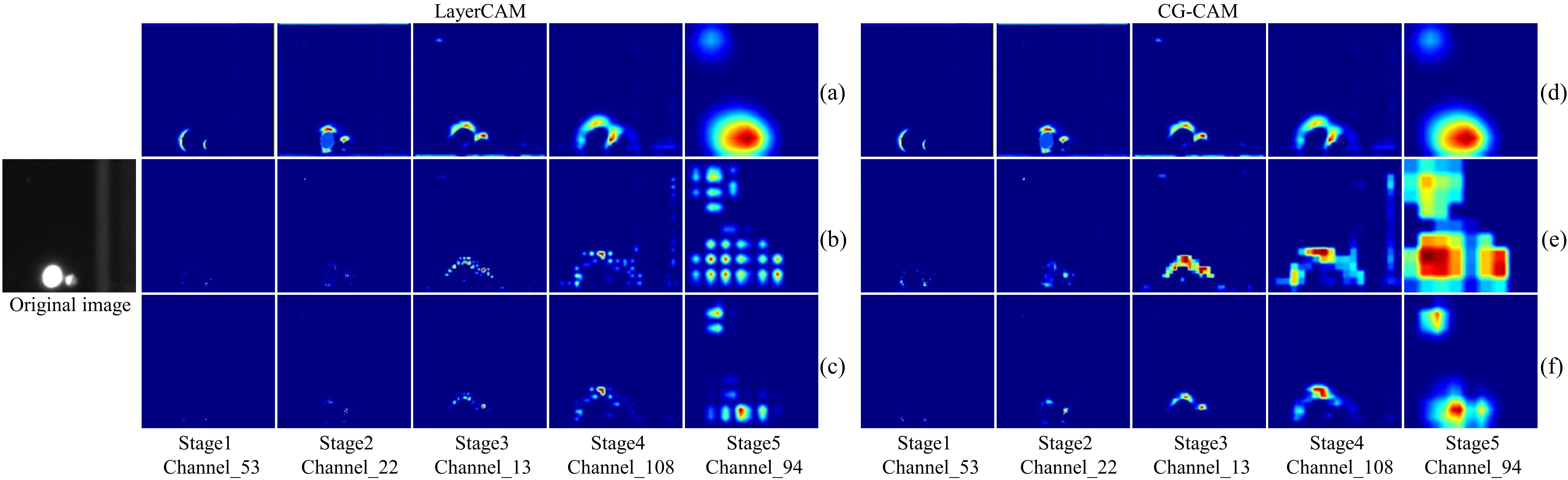}
\caption{Comparison between LayerCAM and CG-CAM from different stages. (a), (d) are the feature maps. (b), (e) are the class activation gradient maps. (c), (f) are the channel class activation maps.}
\end{figure*}

\subsection{Nonlinear multi-scale fusion algorithm}
\paragraph{Analysis} LayerCAM further enhances the activation effect by combining shallow class activation maps, but introduces the problem of under-activation of large targets. As can be seen in Figure 9, small targets are well activated, but large targets are only activated at the edges in the class activation maps from shallow layers. The activation differences can be caused by the characteristics of the network itself. The receptive fields of the shallow layers are relatively small and tend to capture some detailed features, such as the edges and corners of the targets\cite{bib25}. As the depth increases, the receptive field of the network gradually expands, eventually capturing the entire contour of the targets\cite{bib26}.

\begin{figure}[ht]
\centering
\includegraphics[scale=0.39]{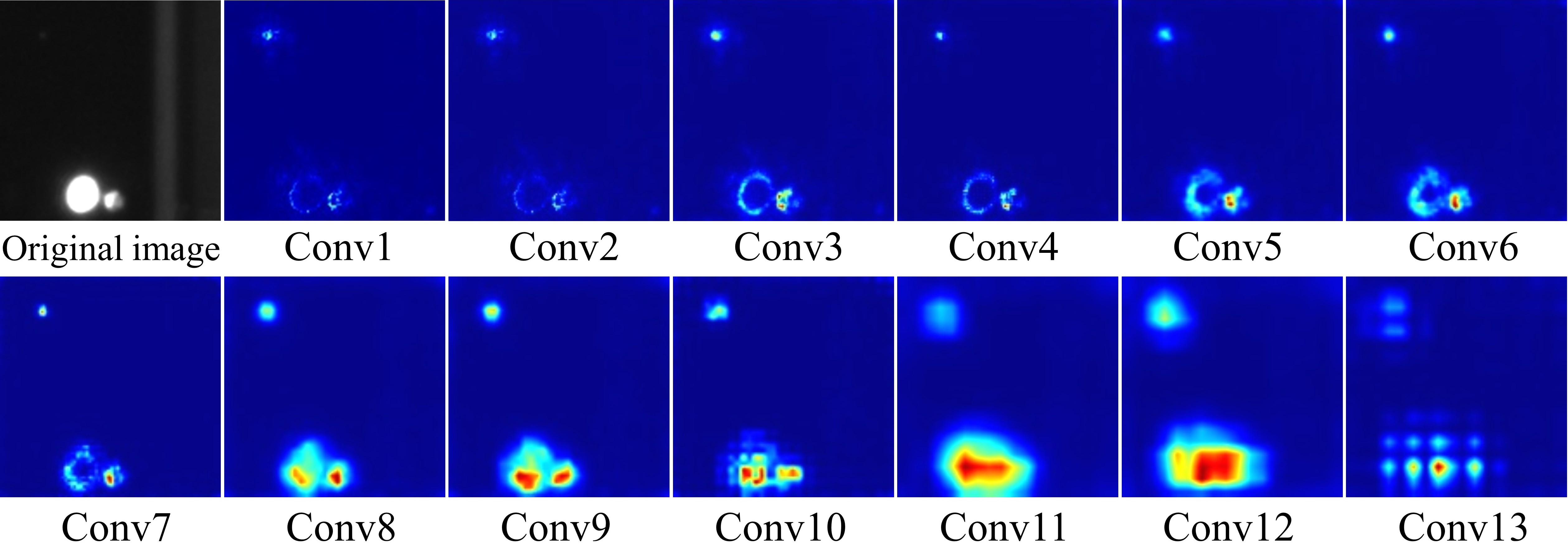}
\caption{Results of LayerCAM from each convolutional layer.}
\end{figure}

Under-activation degrades location accuracy or even leads to missed detections, requiring compensation for under-activated regions of large damage sites. Linearly fusing the over-activated results from deep layers can mitigate this under-activation to some extent, but also masks some of the fine-grained features in shallow layers, resulting in some loss of segmentation accuracy. We find that the gray values of the large targets are low in the class activation maps from the shallow layers, but high in the original images. The opposite is true for small targets. Therefore, we propose to use the original images to compensate for the under-activation of large targets in CG-CAM.

\paragraph{Algorithm} Formally, based on the complementary property of gray values, NM-Fusion adds the gray values of the original image $I$ and those of the class activation map $M_{\text{CG-CAM}}$ to generate a new class activation map $M_{\text{multi}}$. This eliminates the negative impact of under-activation on segmentation, while high-resolution original images do not degrade the fine granularity of $M_{\text{multi}}$. In addition, NM-Fusion extracts the foreground information of the class activation map generated from the final convolutional layer as its mask to filter out the stray light interference introduced by $I$. The class activation map of NM-Fusion $M_{\text{fusion}}$ is calculated as follows:

\begin{equation}
\begin{aligned}
M_{\text {fusion}} & =\left(I+M_{\text {CG-CAM}}\right) \times \text { mask } \\
& =M_{\text {multi}} \times \varepsilon\left(M_{\text {deep }}-v_{\text {thr}}\right),
\end{aligned}
\end{equation}

\begin{equation}
\varepsilon(\text {input})=\left\{\begin{array}{ll}
0 & \text {input} <0 \\
1 & \text {input} \geq 0  
\end{array}\right.,
\end{equation}
Where $v_{\text{thr}}$ is a reasonable threshold. $\varepsilon$() is the step function for generating the mask. The class activation map $M_{\text{deep}}$ of the final convolutional layer has reliable high-level semantic information with a clean and low-noise background. Its over-activated foregrounds can non-linearly activate high-resolution target regions in $M_{\text{multi}}$ without losing fine-grained information. $M_{\text{deep}}$ is calculated as follows:

\begin{equation}
\begin{aligned}
M_{\text {deep }} & =\operatorname{ReLU}\left(\sum_{k} w_{k}^{c} \cdot A_{k}\right) \\
& =\operatorname{ReLU}\left(\sum_{k}\left(\frac{1}{N} \sum_{i} \sum_{j} g_{i j}^{k c}\right) \cdot A_{k}\right),
\end{aligned}
\end{equation}
where N denotes the number of locations in the feature map $A_{k}$. To further optimize the quality of $M_{\text{CG-CAM}}$, we linearly fuse the multi-scale class activation maps from the first 4 stages of CG-CAM. $M_{\text{CG-CAM}}$ is calculated as follows:

\begin{equation}
M_{\text {CG-CAM}}=\sum_{i=1}^{4} M_{\text {Stage } \_i},
\end{equation}
Where, shage\_i is the class activation map of CG-CAM generated from the last convolutional layer at the i-th stage. The results of CG-CAM differ at different stages. As the stage becomes shallower, the class activation maps tend to capture more fine-grained features, but the under-activation problem becomes more severe. We propose to linearly fuse class activation maps from multiple stages to achieve optimal results.

\begin{figure*}[ht]
\centering
\includegraphics[scale=0.57]{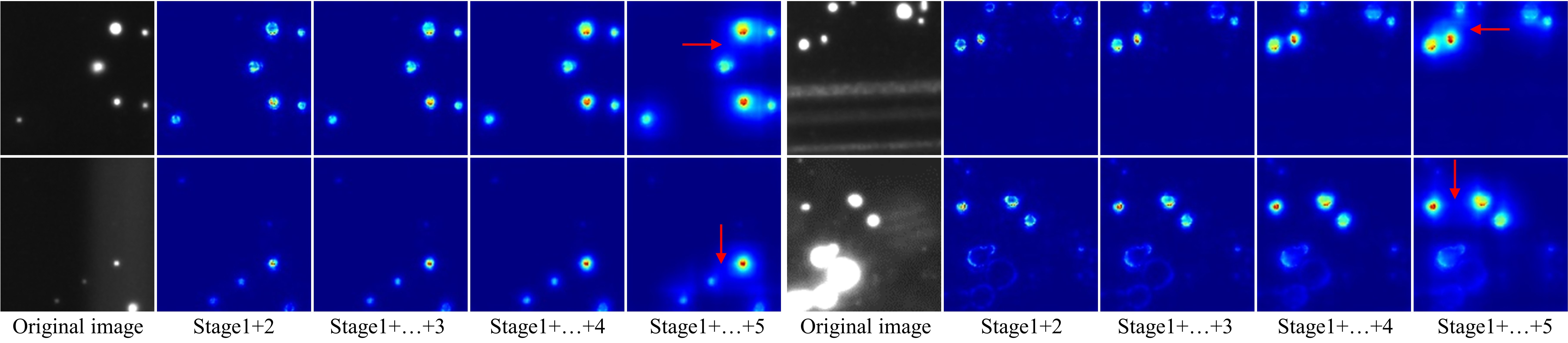}
\caption{Comparison of the multi-scale fusion effect from different stages of CG-CAM}
\end{figure*} 

The comparison of the fusion results from different stages is shown in Figure 10. As the depth of the fusion stage increases, the fusion of the first four stages reaches the best state, with clear boundaries of the target activation regions. However, the addition of the last stage degrades the activation quality and the target boundaries become blurred. This is because the last stage is the deep layer of the network with low resolution of feature maps. Fusing the rough target position information from Stage5 cannot improve the fine granularity. Thus, $M_{\text{CG-CAM}}$ is linear summation of the first 4 stages.

\subsection{Post-processing}
Accurate segmentation of the foreground in the class activation maps is required to achieve precise localization and extraction of damage sites. Simple threshold segmentation cannot achieve ideal segmentation results, so we use the local dynamic threshold segmentation algorithm\cite{bib27} as post-processing. Using a sliding window to iterate through all image regions, an appropriate segmentation threshold $T(i,j)$ is determined based on the contrast of gray values in the local window. It is calculated as follows:

\begin{equation}
T(i, j)=\mu(i, j)\left(1+k\left(\frac{\sigma(i, j)}{R}-1\right)\right),
\end{equation}
Where ($i$,$j$) is the pixel position. $\mu(i, j)$ is the local mean gray value within the window. $\sigma(i, j)$ denotes the corresponding standard deviation. $R$ is the assumed maximum value of the standard deviation. $k$ is the sensitivity parameter. By setting appropriate parameters, The local dynamic threshold segmentation algorithm can adaptively segment damage sites in $M_{\text{fusion}}$.

\section{Experiments}
\subsection{FOA dataset}
We acquired large-aperture images of FOAs in the CAEP high-power laser facility online using the FODI system, and produced the FOA damage dataset through cropping processing and dataset enhancement.

\paragraph{Cropping processing}
The resolution of large-aperture images is 4096$\times$4096, which is too large for the input of the neural network. We crop the images using a 128×128 sliding window with a step size of 64. The 50\% window overlap ensures coherence of information between adjacent windows.

\paragraph{Dataset enhancement}
To improve the ability of the classification network to extract effective features that distinguish between stray light interference and damage sites, we need to enhance the dataset. Due to the limited occurrence of some stray light interference and large damage sites in ICF experiments, it is difficult to collect large amounts of samples. Therefore, we artificially superimpose the damage image on the stray light interference image to produce a new damage image. Figure 11 shows the typical damage class samples, background class samples and manually produced damage class samples.

\begin{figure}[ht]
\centering
\includegraphics[scale=0.44]{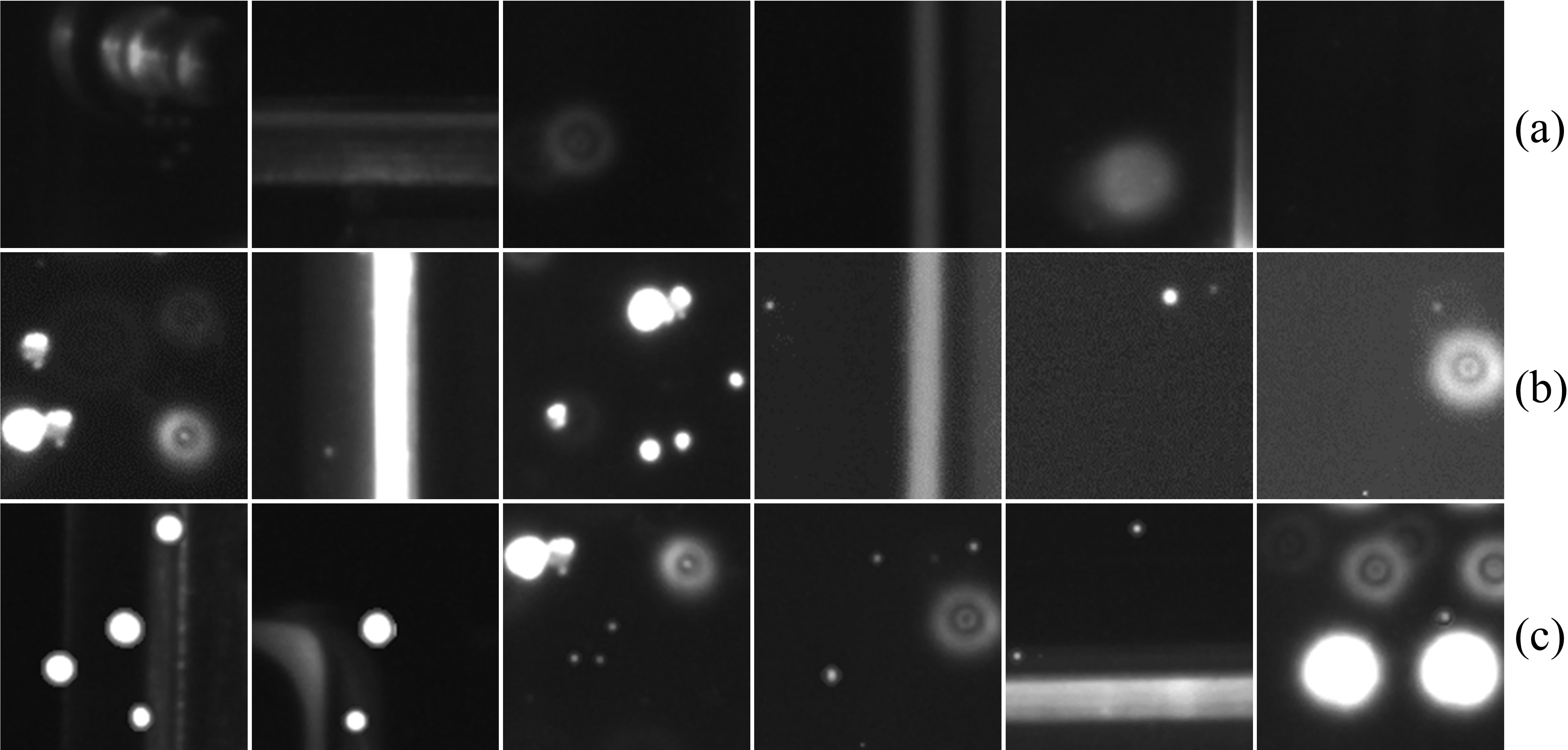}
\caption{Examples of typical samples. (a) Damage class samples. (b) Background class samples. (c) Manually produced damage class samples.}
\end{figure}

The FOA damage dataset we produced contains 1155 training samples and 512 test samples. The ratio of damage class samples to background class samples is approximately 1:1. The training samples include 175 manually produced damage class images and 118 background class images with various stray light interference selected for dataset enhancement. To objectively reflect the performance of the classification network, the test samples cover all types of stray light interference and damage sites.

\subsection{Evaluation Metrics}
In the classification task, Accuracy, Precision, Recall, F1, and False Positive Rate (FPR) are used to evaluate the classification performance of the network (calculated from the damage class, excluding the background class). The statistical objects are images.

In the semantic segmentation task, we provide pixel Recall (p-Recall), pixel Precision(p-Precision), pixel F1(p-F1) and pixel Intersection over Union (IoU) to evaluate the performance of the algorithms. The statistical objects are pixels. The pixel Intersection over Union (IoU) is defined as

\begin{equation}
\text {IoU}=\frac{\text { groud-truth pixels } \cap \text { predicted pixels }}{\text { groud-truth pixels } \bigcup \text { predicted pixels }}.
\end{equation}

In addition, when faced with multiple targets, pixel-level evaluation metrics can mask the negative impact of incorrectly detecting small targets, resulting in a large number of false-positive regions in the results. Therefore, we additionally provide a target-level metric, False Detection Rate (FDR), to evaluate the target detection performance of the algorithm. It is defined as

\begin{equation}
\text {FDR}=\frac{\text {FP}}{\text {TP}+\text {FP}},
\end{equation}
Where TP is the number of objects correctly detected as damage sites. FP is the number of objects incorrectly detected as damage sites. Referring to the definition of evaluation metrics in the object localization task\cite{bib28}, we use pixel IoU to determine whether each detected connected domain is a real damage site or not. The determining formula is as follows:

\begin{equation}
\left\{\begin{array}{l}
\text {Predicted target region}=1, \text {if}( \text{IoU} \geq \delta) \\
\text {Predicted target region}=0,  \text {if}(\text {IoU}<\delta).
\end{array}\right.
\end{equation}

In this paper, $\delta$ is set to 0 (usually set to 0.5) in order to compare the false detection performance of the algorithms. Otherwise GradCAM cannot reasonably calculate this metric due to low IoU values caused by rough segmentation results.

\subsection{Classification experiment}

\paragraph{Experimental Settings}In the experiments, only image-level labels are used to train the VGG-16 classification network. Each sample is flipped horizontally and vertically with a 50\% probability before being fed into the network to increase the generalization ability of the network. The initial parameters are loaded with weights pre-trained on ImageNet to avoid local optima or saddle points and to allow the network to converge quickly. The initial learning rate is set to $1e^{-3}$ and a decay strategy is implemented.The batch size is set to 32 and the training epoch is set to 30 for iterative training. The optimiser in this paper uses the SGD stochastic gradient descent optimiser to update the network weights.

In Table 1, we report the classification performance of the VGG-16 model trained on the enhanced dataset. It can be seen that the F1 of the classification network reaches 97.75\%, with the ability to effectively identify damage sites. 

\begin{table}[ht]
\centering
\caption{The classification performance of the VGG-16 model.}
\scalebox{0.87}{
\begin{tabular}{cccccc}
\hline
Model & Accuracy & Precision &	Recall & FPR & F1\\
\hline
VGG-16 & 97.54\% & 97.13\% &98.39\%  & 3.49\%	& 97.75\% \\
\hline
\end{tabular}
}
\end{table}

\subsection{Weakly-supervised segmentation}
To illustrate the semantic segmentation performance of our method, we selected 338 damage class images containing typical stray light interference as the test set for semantic segmentation and target detection, and manually produced their pixel-level labels (ground truth). We compare the semantic segmentation and target detection results of five baseline algorithms, as shown in Table 2 and Figure 12. Among them, LASNR belongs to the conventional methods. VGG16-Unet and DeepLabv3 (Resnet50 as backbone) belong to the fully supervised semantic segmentation methods. Grad-CAM and LayerCAM belong to the weakly supervised semantic segmentation methods.

\begin{figure*}[ht]
\centering
\includegraphics[scale=0.67]{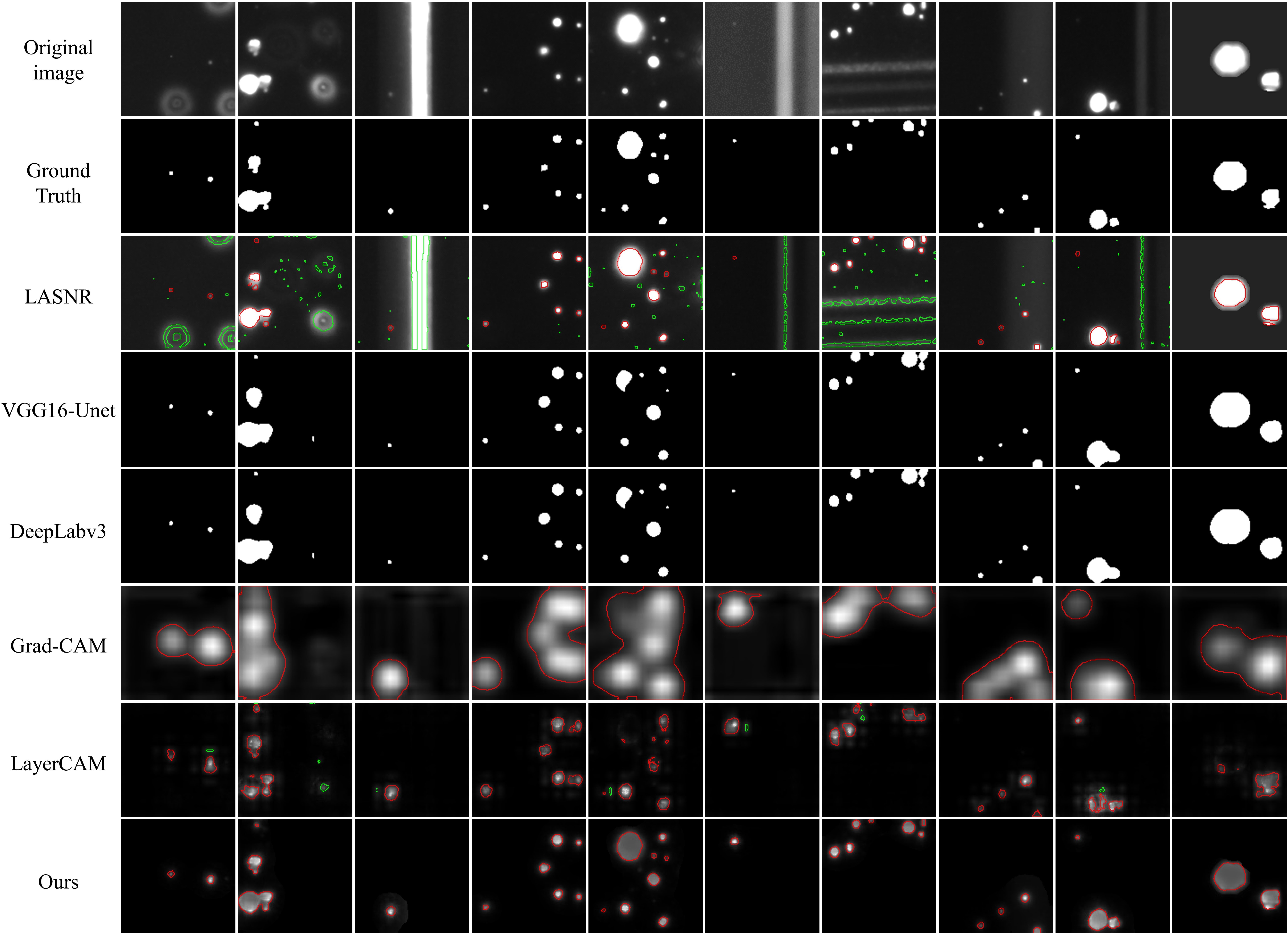}
\caption{Comparison of class activation maps and segmentation results between the baselines and our method. The green areas are the false positive segmentation results. The red areas are the segmentation results containing true damage sites.}
\end{figure*}

\begin{table}[ht]
\centering
\caption{Comparison of baselines and our method under various evaluation metrics.}
\scalebox{0.81}{
\begin{tabular}{cccccc}
\hline
Methods & p-P(\%) & p-R(\%) & p-F1(\%) & t-FDR(\%) & IoU(\%) \\
\hline
LASNR & 63.39 & 99.27 & 70.18  & 41.49 & 37.21 \\
VGG16-Unet & 76.34 & 86.53 & 81.11 & 3.05 & 63.87\\
DeepLabv3 & 81.34 & 86.50 & 83.84 & 3.20& 68.32\\
Grad-CAM&	7.12  & 97.95  & 12.78  &	5.51  & 7.10   \\
LaryerCAM &	61.90  & 89.73  & 	70.89	& 19.57 &	41.82 \\ 
ours &	84.24 &	93.55 &	87.32 &	0.90 &	63.78 \\
\hline
\end{tabular}
}
\tabnote{\textit{Notes:} p-P is an abbreviation for p-Precision. p-R is an abbreviation for p-Recall.}
\end{table}

The experimental results showed that LASNR has a strong segmentation capability with a pixel Recall of 99.27\%, but produces high false detection results. In comparison, Grad-CAM has a much lower FDR of 5.51\%, but has rough class activation regions that are unable to segment multiple targets into independent individuals. The segmentation results from LayerCAM are much finer with an IoU of 41.82\%, which is better than the results from the two baselines above. However, the under-activation of LayerCAM leads to missed detection of large damage sites with a Recall of less than 90\%. In addition, its scattered activation regions result in a high FDR of 19.57\%.

Our method produces more fine-grained class activation maps, and the class activation regions are activated appropriately for various sizes of damage sites. Our IoU metric is up to 63.78\%, an improvement of 56.68\% and 21.96\% over Grad-CAM and LayerCAM respectively, which is comparable to that of fully supervised segmentation algorithms. In addition, our method can suppress the stray light interference with the lowest FDR of all algorithms, only 0.90\%.

\subsection{Ablation experiment}
We successively remove two algorithms from the pipeline: CG-CAM and NM-Fusion, and test the detection and segmentation performance of the remaining algorithms. This experiment demonstrates their respective effect on optimising class activation maps and improving damage segmentation performance. The results are shown in Table 3, Figure 13 and Figure 14.

\begin{figure}[ht]
\centering
\includegraphics[scale=0.45]{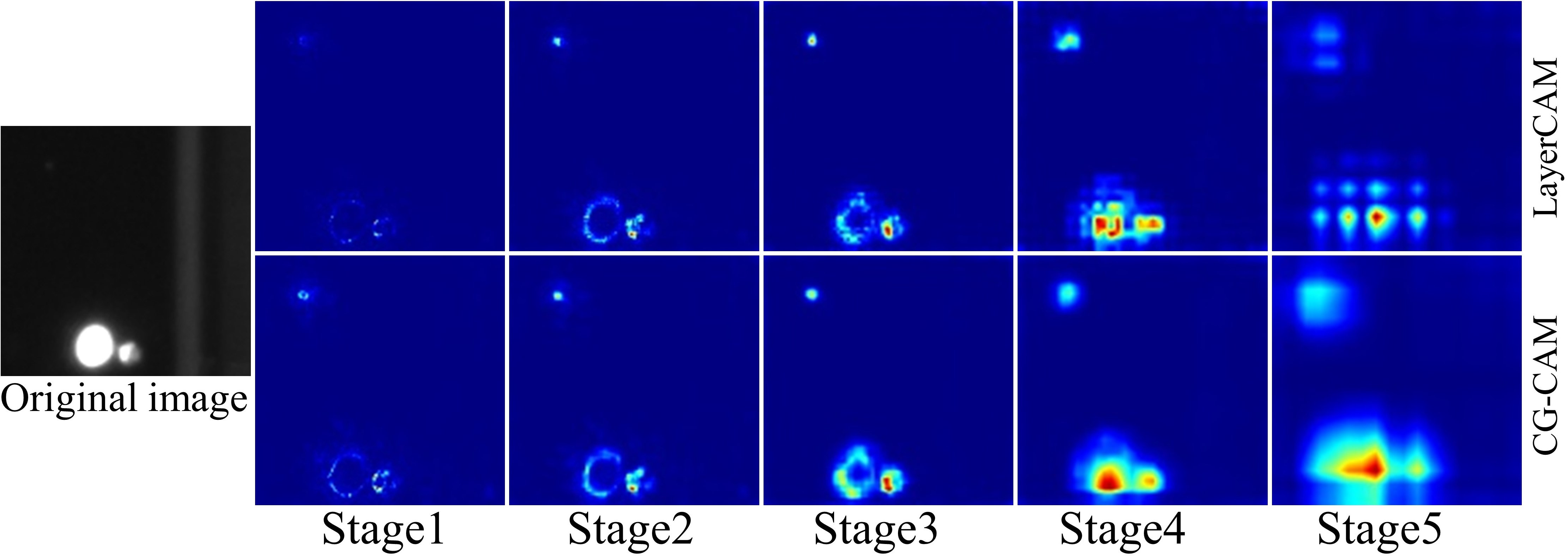}
\caption{Comparison of the class activation maps between LayerCAM and CG-CAM from different stages.}
\end{figure}

\begin{table}[ht]
\centering
\caption{Comparison of the baseline and our two core algorithms under various evaluation metrics.}
\scalebox{0.83}{
\begin{tabular}{lccccc}
\hline
Methods	& p-P(\%) &	p-R(\%)& p-F1(\%)& t-FDR(\%) &IoU(\%)\\
\hline
LaryerCAM & 61.90  & 89.73  &70.89	&19.57 &41.82 \\ 
CG-CAM &75.52&94.87 &	82.53 &8.68 &52.17 \\
\multirow{2}{*}{\makecell{CG-CAM+\\NM-Fusion}}& \multirow{2}{*}{84.24}& \multirow{2}{*}{93.55} &\multirow{2}{*}{87.32} &\multirow{2}{*}{0.90} &\multirow{2}{*}{63.78} \\
& & & &\\
\hline
\end{tabular}
}
\tabnote{\textit{Notes:} p-P is an abbreviation for p-Precision. p-R is an abbreviation for p-Recall.}
\end{table}

CG-CAM restores the continuity of the gradients, preserves high-resolution features with rich detail information, and allows scattered class activation regions to form a whole with semantic meaning (Figure 13). The IoU of CG-CAM is improved by 10.35\% and the FDR is reduced by 10.89\% compared to LayerCAM.

Based on the CG-CAM results, the effect of each step in NM-Fusion is shown in Figure 14. Compared to the results from a single stage, the fusion of multi-scale heatmaps from shallow layers further improves the spatial resolution of class activation maps. Compensated by the original images, the under-activation of large damage sites is effectively improved. Finally, semantic class activation masks from the final convolutional layers non-linearly filter out the stray light interference to produce more fine-grained class activation maps with appropriate activation degree. After the addition of NM-Fusion, the IoU is improved by 11.61\% compared to CG-CAM and the FDR of the damage sites is significantly reduced by 7.78\%.

\begin{figure*}[ht]
\centering
\includegraphics[scale=0.66]{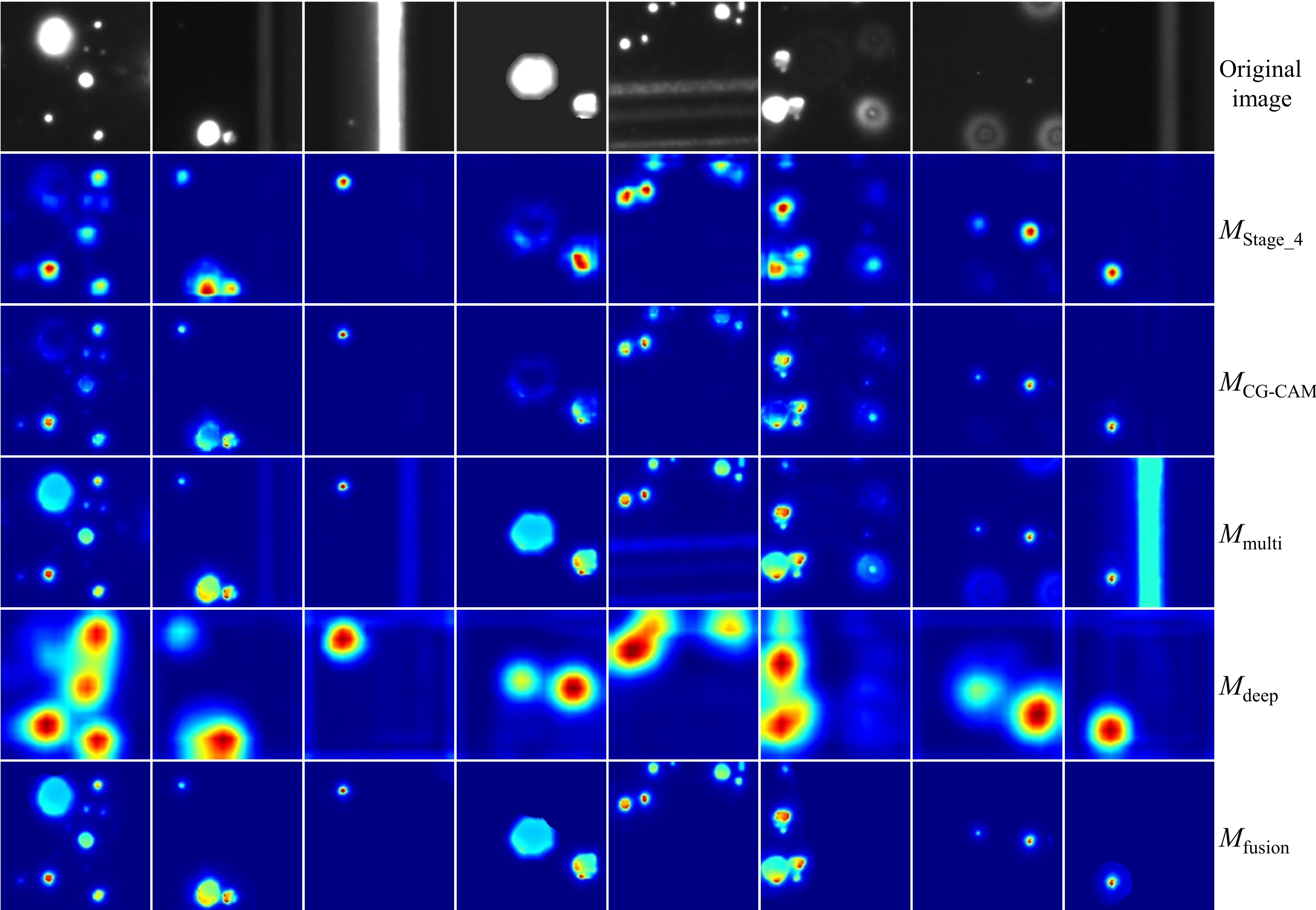}
\caption{The effect of each step in the nonlinear multi-scale fusion algorithm.}
\end{figure*}

\section{Conclusions}
In this paper, a weakly supervised semantic segmentation method with continuous gradient CAM and its nonlinear multi-scale fusion has been proposed for online segmentation of laser-induced damage on large-aperture optics in ICF facilities in the face of complicated damage morphology, uneven illumination and stray light interference. And the classification, detection, and segmentation performance of our method has been tested on the FOA damage dataset. Experiments show that our method can generate appropriately activated high-resolution class activation maps for damage targets of various sizes, with better target detection and segmentation capabilities than current CAM methods. Relying only on image-level labels and limited sample training, our method has achieved segmentation performance comparable to that of fully supervised algorithms, with an IoU of 63.78\%. False detection of damage sites has also been effectively suppressed, with an FDR of 0.90\%. The experimental results demonstrate the effectiveness of the method in this paper. The entire damage segmentation method has been applied to the large laser facilities, giving the FODI system the ability to detect damage online. In the future, we will delve deeper into the relationship between feature maps and class activation gradients to further improve the performance of weakly supervised semantic segmentation.


\end{document}